\documentclass[letterpaper, 10 pt, comsoc]{ieeeconf}
\usepackage[utf8]{inputenc}
\IEEEoverridecommandlockouts %

\usepackage[style=ieee, backend=biber, maxbibnames=5, mincitenames=1, maxcitenames=2, url=false, doi=false, isbn=false]{biblatex}

\usepackage{todonotes}

\usepackage{glossaries}
\usepackage[hidelinks]{hyperref}
\usepackage{booktabs}

\usepackage{cleveref}
\usepackage{float}
\usepackage{placeins}

\newcommand{\tableref}[1]{Table~\ref{#1}}
\newcommand{\figref}[1]{Fig.~\ref{#1}}
\newcommand{\equationref}[1]{Eq.~\ref{#1}}
\newcommand{\secref}[1]{Section~\ref{#1}}

\newcommand{\orcidID}[1]{\href{https://orcid.org/#1}{\includegraphics[width=1em]{include/ORCID-iD_icon-64x64.png}}}

\renewcommand{\orcidID}[1]{}

\newcommand\copyrighttext{%
	\scriptsize \textcolor{blue}{\textcopyright 2019 IEEE. Personal use of this material is permitted.  Permission from IEEE must be obtained for all other uses, in any current or future media, including reprinting/republishing this material for advertising or promotional purposes, creating new collective works, for resale or redistribution to servers or lists, or reuse of any copyrighted component of this work in other works}}
\newcommand\copyrightnotice{%
	\begin{tikzpicture}[remember picture,overlay]
	\node[anchor=north,yshift=-7.5pt] at (current page.north) {\fbox{\parbox{\dimexpr\textwidth-\fboxsep-\fboxrule\relax}{\copyrighttext}}};
	\end{tikzpicture}%
}

\crefrangelabelformat{section}{#3#1#4--#5\crefstripprefix{#1}{#2}#6}
\newacronym{GCN}{GCN}{Graph Convolutional Network}
\newacronym{GNN}{GNN}{Graph Neural Network}
\newacronym{GAT}{GAT}{Graph Attention Network}
\newacronym{CNN}{CNN}{Convolutional Neural Network}
\newacronym{RNN}{RNN}{Recurrent Neural Network}
\newacronym{CVM}{CVM}{Constant Velocity Model}
\newacronym{IDM}{IDM}{Intelligent Driver Model}

\newacronym{HMM}{HMM}{Hidden Markov Model}

   \usepackage{ifthen}
   
   \makeatletter
   \newcounter{IEEE@bibentries}
   \renewcommand\IEEEtriggeratref[1]{%
     \renewbibmacro{finentry}{%
       \stepcounter{IEEE@bibentries}%
       \ifthenelse{\equal{\value{IEEE@bibentries}}{#1}}
       {\finentry\@IEEEtriggercmd}
       {\finentry}%
     }%
   }
   \makeatother

\begin{document}
\title{Graph Neural Networks for Modelling Traffic Participant Interaction}
\author{Frederik Diehl$^{\dagger}$ \orcidID{0000-0002-2421-1801} and Thomas Brunner$^{\dagger}$ \orcidID{0000-0003-0384-5132} and Michael Truong Le$^{\dagger}$ and Alois Knoll$^{\ddagger}$%
	\thanks{$^{\dagger}$Frederik Diehl, Thomas Brunner, and Michael Truong Le are with fortiss GmbH, affiliated institute of Technische Universit\"{a}t M\"{u}nchen, Munich, Germany}%
	\thanks{$^{\ddagger}$Alois Knoll is with the Chair of Robotics, Artificial Intelligence and Real-time Systems, Technische Universit\"{a}t M\"{u}nchen, Munich, Germany}%
}

\maketitle

\copyrightnotice

\global\csname @topnum\endcsname 0
\global\csname @botnum\endcsname 0

\begin{abstract}
By interpreting a traffic scene as a graph of interacting vehicles, we gain a flexible abstract representation which allows us to apply \gls{GNN} models for traffic prediction.
These naturally take interaction between traffic participants into account while being computationally efficient and providing large model capacity.
We evaluate two state-of-the art \gls{GNN} architectures and introduce several adaptations for our specific scenario.
We show that prediction error in scenarios with much interaction decreases by 30\% compared to a model that does not take interactions into account.
This suggests that interaction is important, and shows that we can model it using graphs. This makes \glspl{GNN} a worthwhile addition to traffic prediction systems.
\end{abstract}

\IEEEpeerreviewmaketitle

\section{Introduction}

Short-term accurate behavior prediction of traffic participants is important for applications such as automated driving or infrastructure-assisted human driving\parencite{hinz_designing_2017}. A major open research question is how to model \textit{interaction} between traffic participants. In the past, interactions have been modelled by either creating a representation of one or several traffic participants \parencite{treiber_congested_2000, lenz_deep_2017} or by using a fixed environment representation such as a simulated lidar beam~\parencite{kuefler_imitating_2017}.

However, these methods impose certain disadvantages: A fixed environment representation poses a much harder problem to learn, since we cannot use data we might have extracted previously. Traffic participant representations, on the other hand, scale computationally with the amount of possible interactions, require a human to decide on a useful representation, and underspecify the problem one should learn.

\begin{figure}
	\centering
	\includegraphics[width=1.0\linewidth]{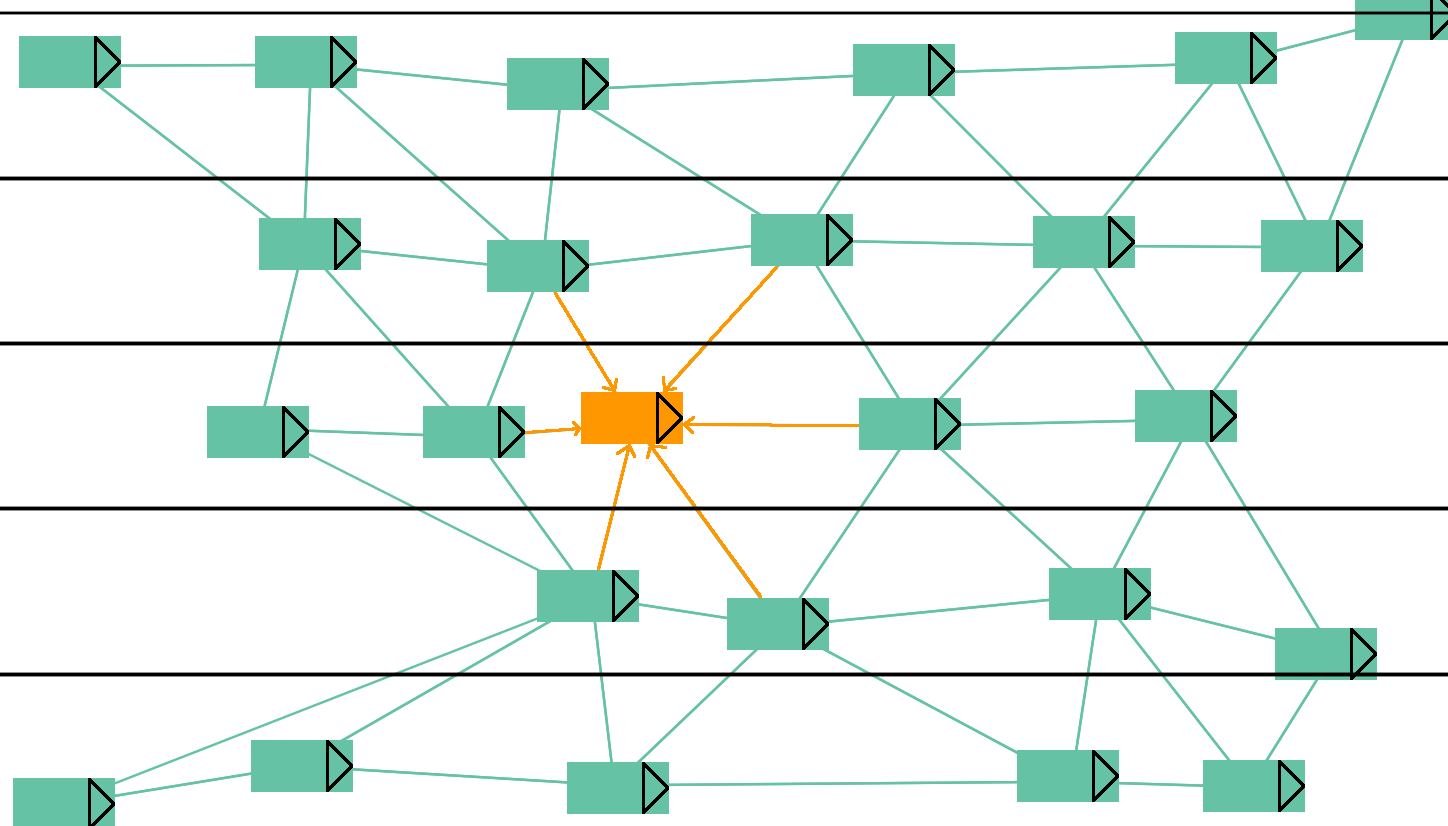}
	\caption{Interaction graph of a traffic situation. Interactions are assumed to occur between the ego vehicle (one vehicle and connections from the vehicles it interacts with are highlighted in orange, but all vehicles are processed simultaneously) and its up to eight neighbours, assigned by current lane (black lines separate lanes). These interactions are modelled as directed edges leading from the neighbouring vehicles to the ego vehicle. Except for the example ego vehicle, edge directions have been omitted for clarity. As can  be seen, this representation is sparse and models the whole traffic situation at once. Best viewed in color.}
	\label{fig:teaser}
\end{figure}

By modelling each vehicle a node and possible interactions between vehicles as edges (see \figref{fig:teaser} for a visualization), we gain a sparse and high-level representation of a traffic scene as a graph.

At the same time, it has been shown \parencite{Morton2016, kuefler_imitating_2017, lenz_deep_2017} that machine learning models and particularly (deep) neural networks perform well on this problem. Yet most available deep learning models operate on data of a fixed size and with a fixed spatial organization such as single data points, time series, or images.

Only fairly recently \parencite{gori_new_2005, scarselli_graph_2009} have \glspl{GNN}, i.e. neural networks operating on graph data, seen research interest and enjoyed successes. Later models \parencite{kipf_semi-supervised_2016, velickovic_graph_2017} only operate on a node's local neighbourhood. This greatly improves scalability while improving performance.

Marrying the representation of a traffic situation as a graph with the modelling capabilities of \gls{GNN} models promises a clear method to take interactions between traffic participants into account, good predictive performance, and efficient computation.

To evaluate this, we conduct traffic participant prediction on two real-world datasets, evaluating their predictive performance and comparing them to three baseline models. We show that prediction error decreases by 30\% compared to our baseline when interaction is plentiful and performs no worse when little interaction occurs. At the same time, computational complexity remains reasonable and scales with linearly in the number of interactions.

This suggests a graph interpretation of interacting traffic participants is a worthwhile addition to traffic prediction systems.

Our main contributions are:

\begin{enumerate}
	\item We show that representing interactions as graphs leads to better performance.
	\item We introduce several adaptations to two state-of-the-art \gls{GNN} models.
	\item We study both the results of different graph construction techniques and our introduced adaptations on two different datasets.
\end{enumerate}

\section{Related Work}
Since traffic participant prediction is a key feature of autonomous driving and traffic simulations, it has been a focus of extensive research for decades. This has lead to a multitude of different algorithms useful for varying prediction timespans and computational resources.

\subsection{Traffic Prediction}

Following the survey by \textcite{Lefevre2014}, we roughly categorize traffic participant prediction into three subgroups of ascending complexity: Physics-based, maneuver-based, and interaction-aware.

\textit{Physics-based models} usually assume little vehicle action and instead use constant velocity or acceleration. The vehicle motion is then predicted from a physical model only. These models can be used for tracking \parencite{Schubert2008} but often fail for predictions longer than a second or when vehicle interaction plays an important role.

\textit{Maneuver-based models} use a set of maneuver prototypes and either match the past trajectory directly using cluster-based approaches \parencite{Vasquez2004} or from vehicle features using machine learning methods \parencite{GarciaOrtiz2011, Morris2011,Kumar2013}. While these models are now able to include more complex maneuver, they also cannot take interaction into account.

\textit{Interaction-aware models} aim to include interactions between vehicles in their predictions. These include an expansion of maneuver-based models which account for collision probabilities \parencite{Lawitzky2013}, coupled \glspl{HMM}, which model pairwise entity dependencies \parencite{Brand1997}, or machine learning-based models. 

Machine learning-based models also vary in complexity and goal. \textcite{lenz_deep_2017} use simple feed-forward neural networks to create a fast model for use in a Monte-Carlo Tree Search algorithm. \textcite{Morton2016} evaluate \glspl{RNN} for the same task, also trained in a supervised fashion. Conversely, \textcite{kuefler_imitating_2017} use Generative Adversarial Imitation Learning to imitate human driving behavior using reinforcement learning. For all of these, performances crucially depend on the representation of the environment.

\subsection{Environmental Representation}
Environmental representation can be differentiated by their abstractness: One can represent the environment as data close to sensor input such as LIDAR beams \parencite{kuefler_imitating_2017}, camera images, or a simple gridmap. Alternatively, one can represent the environment as a list of discrete objects.

\subsubsection{Sensor-like representation} A sensor-like representation does not require expert knowledge to define the features to use and remains of constant size independent of factors such as traffic density. At the same time, it can receive information on many vehicles. However, the representation is inefficient (requiring many LIDAR beams or pixels per vehicle) and we are forced to learn not just driving behavior but also the extraction of vehicles from sensor data.

\subsubsection{Discrete object representation} Alternatively, we can represent each vehicle as an object with certain attributes. Predictions are then created per car. Interaction is then a matter of choosing the correct environment representation, which may be as simple as the distance and approach speed to the preceding vehicle --- as in the \gls{IDM}  \parencite{treiber_congested_2000} --- or might contain a multitude of preprocessed features \parencite{lenz_deep_2017}. However, these models by design have to be simplistic in their assumptions of interaction between traffic participants and are therefore limited.

Several of these shortcomings can be avoided by thinking about traffic participants and their interactions as nodes and edges in a graph. A behavior prediction model then operates on that graph, producing predictions for each node.

\vspace{\baselineskip}

\section{Traffic Participant Prediction from a Graph}

While there are several different \gls{GNN} architectures, experimental results suggest the relatively simple \gls{GCN} model still performs best over a wide variety of tasks \parencite{shchur_pitfalls_2018}. We also evaluate the \gls{GAT} model, since it allows us to easily include edge features into the model.

\subsection{Graph Convolutional Networks}
\glspl{GCN} \parencite{kipf_semi-supervised_2016} are an approach for node-based classification or prediction on a graph. Analogous to convolutions on images or time series, a \gls{GCN} applies the same operation on all nodes. Like other neural networks, it is defined by a series of differently-parameterized layers which are applied successively.

\subsubsection{The Base Model}

Each layer of the \gls{GCN} uses 
\begin{equation}
	H^{(l+1)} = \sigma\left( \tilde{D}^{-\frac{1}{2}} \tilde{A} \tilde{D}^{-\frac{1}{2}} H^{(l)} W^{(l)}\right)
\end{equation}

as a transformation. Here, $H^{(l)}$ is the $l$th layer's activations, $\tilde{A}$ is the adjacency matrix with added self-connections between nodes, $\tilde{D}$ is the degree vector of $\tilde{A}$, and $W^{(l)}$ is the $l$th layer's learnable weight matrix.

This is equivalent to a first-order approximation of a localized spectral filter, but has two crucial advantages: The Graph Laplacian does not need to be inverted (which would incur computational cost of $O(n^3)$) and the transformation specified by $l$ layers takes exactly the $l$-hop neighborhood of a node into account. Accordingly, computational complexity scales linearly in the number of edges and can take vehicles into account which are not directly connected to the ego vehicle. This makes it more efficient than a naive encoding of the neighboring vehicles.

\subsubsection{Adaptations for the \gls{GCN}}

We originally applied the \gls{GCN} exactly as described by \textcite{kipf_semi-supervised_2016}. However, we found several changes to be crucial:

\begin{itemize}
	\item \textit{Residual Weights:} \glspl{GCN} compute the next layer's features for a node from a spectral decomposition of that node's neighborhood (and, with added self-connections, the ego node itself). However, this means a \gls{GCN} cannot treat the ego node's own features differently from any of its neighbors. In the prediction task, this appears to be a significant obstacle to good performance. Accordingly, we remove the self-connections but introduce a second weight matrix defining a transformation on the ego node's features.
	Our transformation equation is therefore
	\begin{equation}
	\small
	H^{(l+1)} = \sigma\left( D^{-\frac{1}{2}} A D^{-\frac{1}{2}} H^{(l)} W^{(l)} + H^{(l)} W^{(l)}_{s}\right).
	\end{equation}
	\item \textit{Weight by Distance:} \textcite{kipf_semi-supervised_2016} note that the adjacency matrix can be binary or weighted. We evaluate weighting edges by the inverse distance, with self-loops set to $1$.
	\item \textit{Feed-forward output:} In addition, we no longer use a full \gls{GCN} but replace the output layer with a feed-forward layer operating on each node's features independently. This allows a better decoupling of the feature extraction (occuring in the first few \gls{GCN} layers) and the final prediction from the extracted features.
\end{itemize}

\subsection{Graph Attention Networks}
\acrfull{GAT} \parencite{velickovic_graph_2017} layers compute each node's next representation by an attention mechanism over all of its neighbors. 

\subsubsection{The Base Model}
Specifically, they compute attention coefficients

\begin{equation}
\label{eq:gat_attention}
	\alpha^l_{ij} = softmax_j \left( a(W^lh^{(l-1)}_i, W^lh^{(l-1)}_j) \right)
\end{equation}

for each connected node pair, with $h^l_i$ being the $i$th node's feature in the $l$th layer and $W^l$ being the learnable weight matrix for the $l$th layer. $a$ is the learnable attention computation, implemented by a neural network. The node feature vector is then computed as 
\begin{equation}
	h^l_i = \sigma \left( \sum_{j \in N_i} \alpha^l_{ij} W^lh^{(l-1)}_j \right),
\end{equation}
where $\sigma$ is a non-linearity, usually ReLU.

In practice, \textcite{velickovic_graph_2017} note that learning is stabilized by using multi-head attention, i.e. using $k$ differently-parameterized attention mechanisms and concatenating - or averaging in the last layer - the result. This allows features to be created from different subsets of nodes depending on the needs of these features.

As with \glspl{GCN}, a \gls{GAT} layer operates on local neighborhood only and therefore also scales linearly in the number of edges.

\subsubsection{Adaptations for the \gls{GAT}}
As before, we also apply some adaptations to the base \gls{GAT} model.

\begin{itemize}
	\item \textit{Edge attributes:} In the \gls{GAT} as introduced by \textcite{velickovic_graph_2017}, attention depends only on the features of the two nodes. However, we do have additional data - like the relative positions - available to us in this scenario. Accordingly, we augment the attention computation from \equationref{eq:gat_attention} by including edge features, such that 
	\begin{equation}
	\alpha^l_{ij} = softmax_j \left( a(W^lh^{(l-1)}_i, W^lh^{(l-1)}_j, e_{ij}) \right).
	\end{equation}
	We do not learn successive edge features but instead use the relative positions for each layer.
	
	\item \textit{Residual Weights:} While the \gls{GAT} should be able to learn by itself to concentrate one attention head onto the ego node, we also evaluate explicitly adding a transformation of the ego node's features.
	\item \textit{Feed-forward output:} As with the \gls{GCN}, our final output is produced by a feed-forward layer.
\end{itemize}

\subsection{Graph and Feature Construction}
\label{sec:graph_construction}
Formulating the prediction problem as a graph still leaves open the task of how we construct said graph and the node features. While there is an obvious strategy to construct node features - namely to use the corresponding car features like position or velocity - no such strategy is apparent to construct connections between the nodes. However, four basic strategies are immediately apparent:

\begin{itemize}
	\item \textit{Self connections:} This only adds self-loops to the graph. It ignores all interaction performance and should perform identically to a simple model operating on the vehicle data only.
	\item \textit{All connections:} Connecting all vehicles ensures that no interactions are ignored. However, this ignores previous knowledge on spatial position and interaction and scales quadratically in the number of vehicles.
	\item \textit{Preceding connection:} Arguably the most important interaction is with the vehicle immediately in front of us. We can therefore construct interactions only between the current vehicle and its predecessor.
	\item \textit{Close vehicles:} Alternatively, we can argue that the main interactions are with the vehicles in an ego vehicle's direct environment, which are at most eight vehicles located to the front, rear, and sides of the ego vehicle. This construction is similar to the approach by \textcite{lenzDeepNeuralNetworks2017a} and \textcite{wheelerFactorGraphScene2016}.
\end{itemize}

While we would prefer to learn these connecting strategies, this is a very difficult open problem and scales quadratically with the number of considered vehicles. We therefore only evaluate the fixed strategies. Connected to this, we also leave the interesting question of what neighborhood size is necessary to future work.

\section{Experiments}

In order to evaluate the newly proposed models, we conduct a prediction experiment on real-world traffic data. We purposely keep baselines and models simple to demonstrate whether the graph interpretation is beneficial without introducing a multitude of confounding factors. We therefore do not include \gls{RNN} architectures, simulation steps, or imitation learning.

From this, we aim to answer three main questions: (A) Which of our adaptions to \glspl{GNN} are necessary? (B) How do we construct an interaction graph? (C) Does a graph model increase prediction quality?

\subsection{Datasets}
We conduct our experiment on two different datasets: The NGSIM I-80 dataset \parencite{NGSIM} and the HighD dataset \parencite{highDdataset}.

\subsubsection{NGSIM}
The NGSIM project's I-80 dataset contains trajectory data for vehicles in a highway merge scenario for three 15-minute timespans. These are tracked using a fixed camera system. As \textcite{Thiemann2008} show, position, velocity, and acceleration data contain unrealistic values. We therefore smooth the positions using double-sided exponential smoothing with a span of 0.5$s$ and compute velocities from these.

We use two of the recordings as the training set and split the last one equally into validation and test set. We subsample the trajectory data to 1 FPS and extract trajectories consisting of a total of 10$s$ of length. The goal of the model is to predict the second half of the trajectory given the first five seconds.

\subsubsection{HighD}
Since the NGSIM dataset still contains many artifacts (errors in bounding boxes, undetected cars, complete non-overlap of bounding box and true vehicle), we additionally conduct experiments on the new HighD dataset \parencite{highDdataset}, which is a series of drone recordings and extracted vehicle features from about 400 meters each from several locations on the German Autobahn. A total of 16.5\,h of data is available, containing 110\,000 vehicles with a total driving distance of 45\,000\,km. However, since the dataset consists mainly of roads without on- or off-ramps and without traffic jams, interaction seems limited: Only about 5\% of the cars experience a lane change.

To avoid information leakage, we split the dataset by recording. The last 10\,\% of the recordings are used as test set, the 10\,\% before that as validation set. Trajectory construction is then identical to the NGSIM dataset.

\subsection{Baselines}

We compare our approach to two different model-based static approaches, and one learned approach.

\subsubsection{\acrfull{CVM}}
This model considers each car to continue moving at the same velocity (both laterally and longitudinally) as the last frame it was observed.

\subsubsection{\acrfull{IDM}}
The \gls{IDM} \parencite{treiber_congested_2000} is a commonly-used driver model for microscopic traffic simulation since it is interpretable and collision-free. We use this to predict the changes in longitudinal velocity and keep the in-lane position constant.

The \gls{IDM}'s acceleration is computed from both a \textit{free road} and an \textit{interaction} term. The free road acceleration is computed as 
\begin{displaymath}
	a_{free} = a_{max} \left( 1 - \left( \frac{v}{v_{0}}\right)^\delta \right),
\end{displaymath}
with the maximum acceleration $a_{max}$, the acceleration exponent $\delta$ and the desired velocity $v_{0}$ being tunable parameters and the current velocity $v$. The interaction term is defined as 
\begin{displaymath}
	a_{int} = -a_{max} \left( \frac{s_0 + v * \tau}{s} + \frac{v \Delta v}{2 s  \sqrt{a_{max} b}}\right),
\end{displaymath}
where the minimum distance to the front vehicle $s_0$, the time gap $\tau$, and the maximum deceleration $b$ are tunable parameters. $v$ is the vehicle's speed and $\Delta v$ the closing speed to its predecessor. The total acceleration is the sum of both the free road and the interaction acceleration. Since the \gls{IDM} only outputs a longitudinal acceleration, we assume no lateral motion when using the \gls{IDM}

We take the \gls{IDM} parameters for the NGSIM dataset from \textcite{Morton2016}. For the HighD dataset, we tune the \gls{IDM}'s parameters using guided random search with a total of 20\,000 samples. Both values are listed in \tableref{tab:idmparams}.

\begin{table}
	\vspace{0.15cm}
	\caption{Optimized parameters of the \gls{IDM}.}
	\label{tab:idmparams}
	\begin{tabular*}{\columnwidth}{l c c @{\extracolsep{\fill}} rr}\toprule
		Parameter & & & HighD & NGSIM \parencite{Morton2016} \\ \midrule
		Desired velocity & $v_0$ & $\left[\frac{m}{s}\right]$ & $58.87$ & $17.8$\\ 
		Maximum acceleration & $a_{max}$ & $\left[\frac{m}{s^2}\right]$& $0.14$ & $0.76$\\ 
		Time gap & $\tau$ & $\left[s\right]$ & $0.12$ & $0.92$\\ 
		Comfortable deceleration & $b$ & $\left[\frac{m}{s^2}\right]$ & $12.17$ & $3.81$\\ 
		Minimum distance & $s$ & $\left[m\right]$ & $14.46$ & $5.249$\\ 
		\bottomrule
	\end{tabular*}
\end{table}

\subsubsection{Independent Feed-Forward Model}
In addition to the models taking interaction into account, we also add a simple feed-forward neural network predicting the trajectory from only the ego vehicle's past data. We use this baseline model to measure the improvement we gain from including interaction into our models.

\subsection{Model Configuration}

Each model uses a similar configuration: Two layers producing a 256-dimensional feature representation followed by a feed-forward layer producing the final output: The displacement in x and y direction. All models use the ReLU nonlinearity. The \gls{GAT} employs four attention heads (and 64-dimensional feature representations each).

Since the \gls{GNN} models use two layers, their effective receptive field is the two-hop neighborhood from the ego vehicle.

All models receive inputs and produce outputs in fixed-length timesteps without recurrence. They are trained to predict displacement relative to the last position and receive position and velocity for each past timestep. They train to minimize the mean squared error over all outputs. All models are implemented in pytorch \parencite{paszke2017automatic} using and expanding upon the pytorch-geometric library \parencite{Fey/etal/2018}.

\subsection{Performance Measure}
We report performances of the model by measuring the error in position between ground truth and prediction. We both report mean displacement over five seconds, weighting each timestep identically, and final displacement after five seconds.

\subsection{Experimental Procedure}
Our choice of experiments is guided by the three main questions (\cref{sec:RQ1,sec:RQ2,sec:RQ3}). To ensure meaningful results, we repeat each evaluation a total of ten times using different, randomly-chosen random seeds. In tables, we report all results as mean $\pm$ standard deviation. Figures are violin plots, showing both individual results and the total result distribution.

We optimize both network adaptations and graph construction strategies on the NGSIM I-80 dataset since it is both smaller and contains more interactions. We then use these insights to pick the best-performing models and evaluate them on both the NGSIM I-80 and the HighD dataset.

\section{Discussion}

\begin{table}
	\vspace{0.15cm}
	\caption{Results for our \gls{GNN} ablation study on the NGSIM I-80 dataset. \newline ($\star$) Uses 3 instead of 10 evaluations.}
	\label{tab:ablation}
	\begin{tabular*}{\columnwidth}{l @{\extracolsep{\fill}} r r}\toprule
		& Mean Displ. $\left[m\right]$ & Displ. @5s $\left[m\right]$ \\ \midrule
		\gls{GCN} Adaptations & & \\ \midrule
		Default     & $2.50 \pm 0.07$ & $4.68 \pm 0.15$\\
		no ff output & $2.52 \pm 0.06$ & $4.66 \pm 0.11$ \\
		with weighted edges & $2.60 \pm 0.08$ & $4.91 \pm 0.15$\\
		no residuals \& weighted edges& $2.87 \pm 0.04$ & $5.19 \pm 0.08$\\
		no residuals & $3.74 \pm 0.11$ & $6.42 \pm 0.10$\\ \midrule
		\gls{GAT} Adaptations & & \\ \midrule
		Default     & $1.92 \pm 0.04$ & $3.45 \pm 0.08$\\
		no ff output & $1.93 \pm 0.07$ & $3.38 \pm 0.16$\\
		no residuals & $2.32 \pm 0.02$ & $3.96 \pm 0.04$\\
		no edge features & $2.40 \pm 0.05$ & $4.48 \pm 0.09$\\\midrule
		Connection Strategy (GAT) & & \\ \midrule	
		Self-Connections     & $2.68 \pm 0.05$ & $5.08 \pm 0.08$\\
		Preceding Connection & $2.70 \pm 0.04$ & $5.11 \pm 0.07$\\
		Neighbour Connection & $1.93 \pm 0.08$ & $3.47 \pm 0.13$\\
		All Connections ($\star$)      & $2.41 \pm 0.02$ & $4.42 \pm 0.03$\\
		\bottomrule
	\end{tabular*}
\end{table}

We structure our evaluation according to three research questions which answer (A) whether our proposed architectural adaptions are worthwhile, (B) which of the graph construction strategies should be preferred, and (C) whether the inclusion of interaction graph information improves performance.

\subsection{Which of our adaptions to \glspl{GNN} are necessary?}
\label{sec:RQ1}

In \secref{sec:graph_construction}, we proposed several changes to the \gls{GCN} and \gls{GAT} architectures. To answer which of these changes are beneficial, we conducted an ablation study whose results are listed in \tableref{tab:ablation}.  We evaluated this using the Neighbour Connection graph construction strategy.

For both models, removing the residual connections decreases the prediction error by at least 20\%. This is likely because there is a clear difference between a neighbouring and the ego node in this task. We also found that using a feed-forward layer as last layer does produce a small increase in performance but also stabilizes training.

Introducing relative positions as edge features into the \gls{GAT} seems to be a clear success, reducing the final displacement by about a meter. Contrary to that, edge weights for the \gls{GCN} slightly decrease performance, especially when omitting residual weights. We believe that the main contribution of edge weights in our scenario is to discern between the ego and surrounding vehicles, which is already more effectively modelled through residual weights.

We therefore evaluate the graph construction using the \gls{GAT} model.

\begin{figure*}
	\vspace{0.15cm}
	\centering
	\begin{minipage}[b]{.49\textwidth}
		\includegraphics[width=1.0\linewidth]{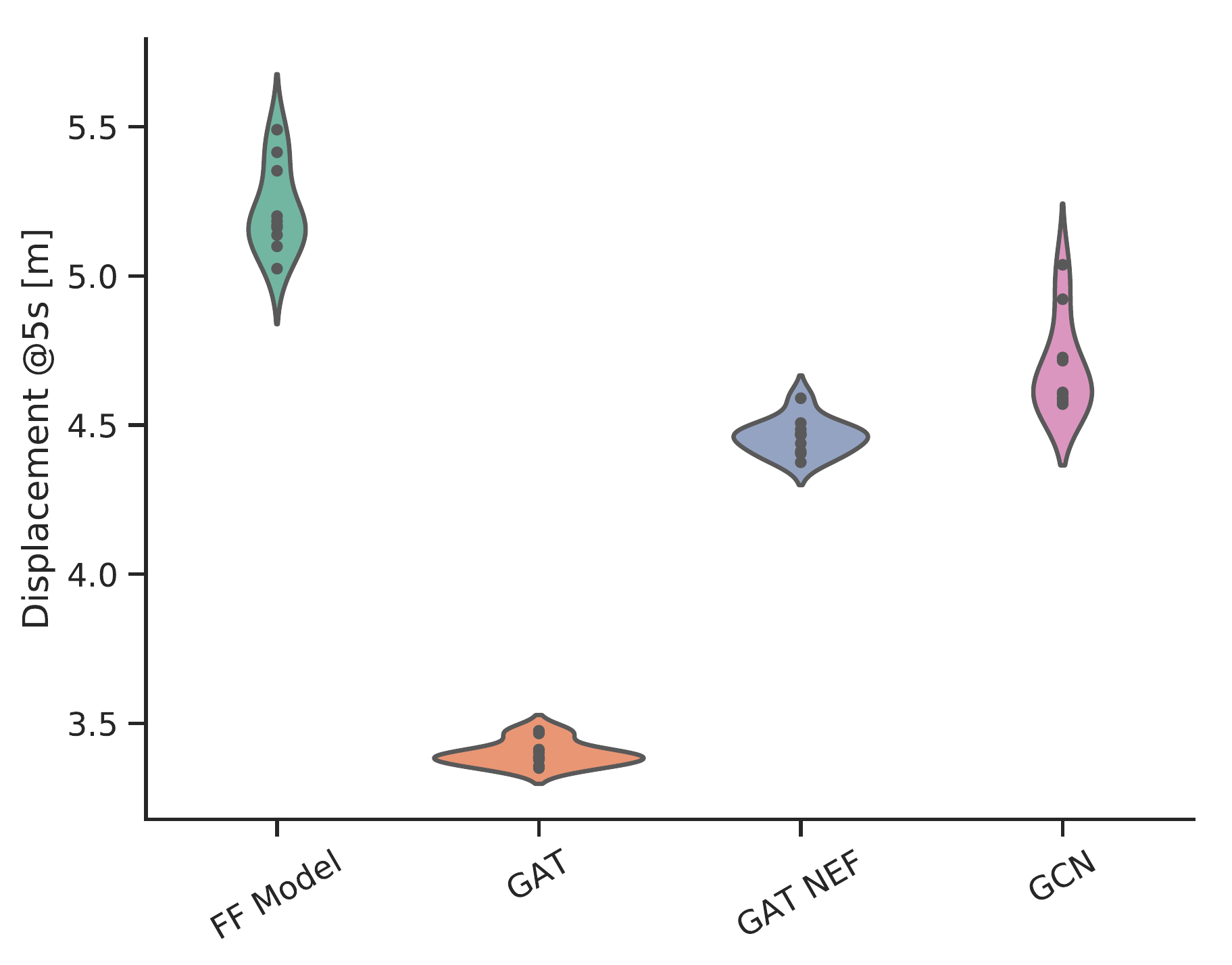}
		\caption{Performance on the NGSIM Dataset. As can be seen, all three \gls{GNN} models 	perform better than the baseline which does not take interaction into account.}
		\label{fig:ngsim_general_mae}
	\end{minipage}\hfill
	\begin{minipage}[b]{.49\textwidth}
		\includegraphics[width=1.0\linewidth]{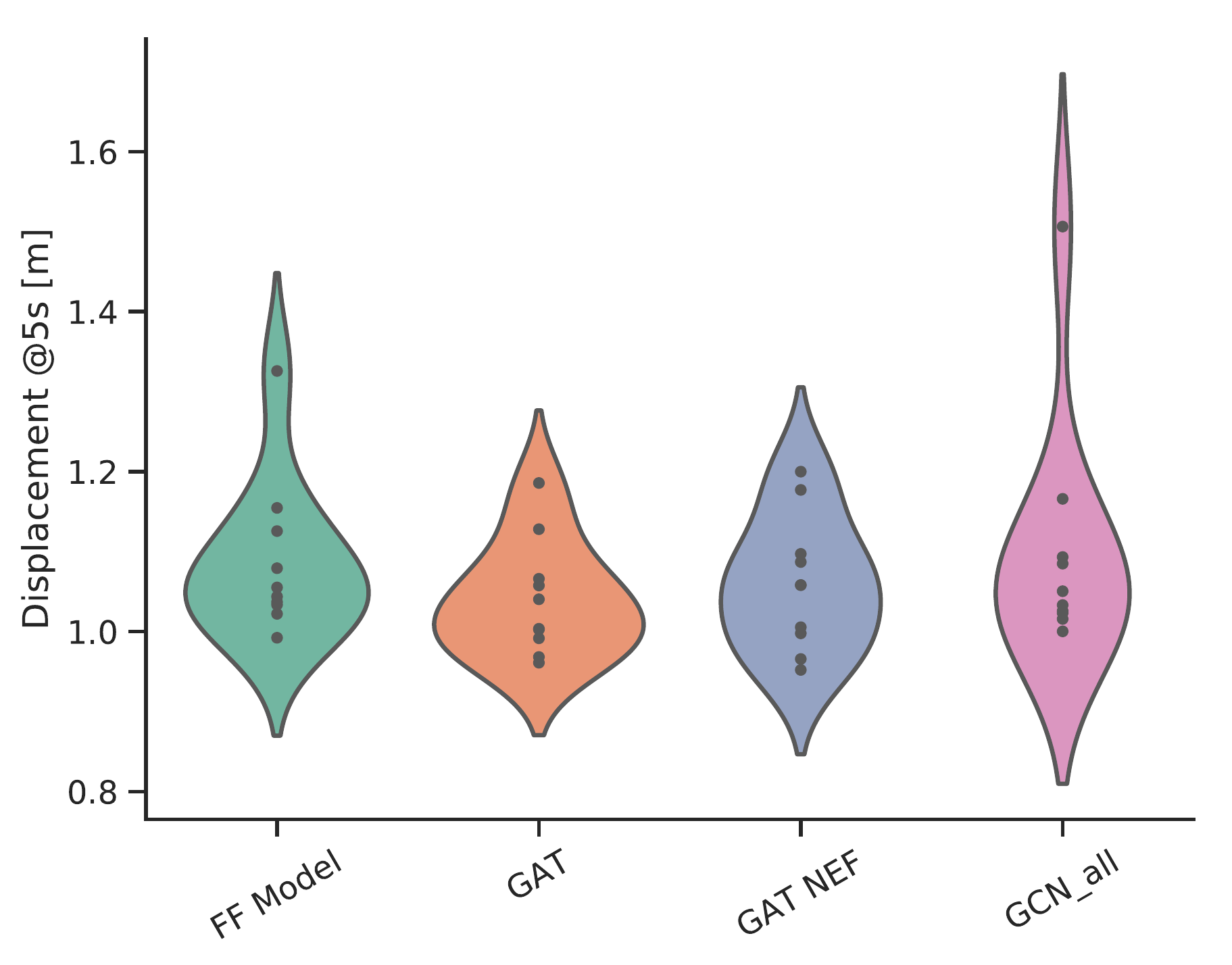}
		\caption{Performance on the HighD Dataset. As can be seen, the algorithms perform similar. We believe this to be due little interaction occurring in the dataset.}
		\label{fig:highd_general_mae}
	\end{minipage}
\end{figure*}

\begin{table*}
	\vspace{0.15cm}
	\centering
	\begin{minipage}[b]{.49\textwidth}
		\caption{Performance comparison on the NGSIM dataset.}
		\label{tab:ngsim_performance}
		\begin{tabular*}{\columnwidth}{l @{\extracolsep{\fill}} r r}\toprule
			& Mean Displ. $\left[m\right]$ & Displ. @5s $\left[m\right]$ \\ \midrule
			GAT     & $1.89 \pm 0.02$ & $3.40 \pm 0.04$\\
			GAT NEF & $2.39 \pm 0.03$ & $4.46 \pm 0.06$\\
			GCN & $2.51 \pm 0.07$ & $4.69 \pm 0.12$\\
			FF & $2.78 \pm 0.08$ & $5.22 \pm 0.14$\\
			CVM      & $2.58 \phantom{\pm 00.14}$ & $5.00 \phantom{\pm 00.14}$\\
			IDM      & $3.10 \phantom{\pm 00.14}$ & $6.60 \phantom{\pm 00.14}$\\\bottomrule
		\end{tabular*}
	\end{minipage}\hfill
	\begin{minipage}[b]{.49\textwidth}
		\caption{Performance comparison on the HighD dataset.}
		\label{tab:highd_performance}
		\begin{tabular*}{\columnwidth}{l @{\extracolsep{\fill}} r r}\toprule
			& Mean Displ. $\left[m\right]$ & Displ. @5s $\left[m\right]$ \\ \midrule
			GAT     & $0.47 \pm 0.04$ & $1.04 \pm 0.07$\\
			GAT NEF & $0.49 \pm 0.07$ & $1.06 \pm 0.08$\\
			GCN      & $0.47 \pm 0.08$ & $1.10 \pm 0.14$\\ 
			FF & $0.45 \pm 0.06$ & $1.09 \pm 0.09$\\
			CVM      & $1.09 \phantom{\pm 00.14}$ & $2.66 \phantom{\pm 00.14}$\\
			IDM      & $1.12 \phantom{\pm 00.14}$ & $2.66 \phantom{\pm 00.14}$\\\bottomrule
		\end{tabular*}
	\end{minipage}
\end{table*}

\subsection{How do we construct an interaction graph?}
\label{sec:RQ2}
In \secref{sec:graph_construction}, we proposed four construction strategies for the interaction graph. We evaluate the quality of predictions with each of these strategies using the \gls{GAT} models, since these seemed to perform best. We note that in practical scenarios, a tradeoff might be necessary between prediction quality and computational complexity. \tableref{tab:ablation} shows results. 

As expected, the Self-Connection strategy performs identically to the FF baseline model, and the Neighbour-Connection graph construction method performs best. Somewhat surprisingly, the Preceding-Connection strategy performs no better than the baseline. 

We especially note that using the All-Connection strategy imposes signficant computational disadvantages with quadratic instead of linear runtime and, in our experiments, a slowdown of about 50x.

We therefore use the Neighbour Connection graph construction strategy for our evaluation.

\subsection{Does a graph model increase prediction quality?}
\label{sec:RQ3}

The motivation of our work is to evaluate whether it is beneficial to model interaction between traffic participants and whether this can be modelled in a graph construction. To answer this question, we compare models with interaction to a model without (FF). We also include a comparison with two classical models (\acrshort{CVM} and \gls{IDM}).

We chose the \gls{GAT} model as best-performing \gls{GNN}. We also included a \gls{GAT} model without edge features (called GAT NEF in our figures and tables) for a fair comparison with the \gls{GCN} model.

\subsubsection{NGSIM} 
\figref{fig:ngsim_general_mae} and \tableref{tab:ngsim_performance} show the performance of both our learning baseline and three of our \gls{GNN} models. 

As can be clearly seen, every \gls{GNN} model performs better than the baseline. At the same time, there are clear performance differences between them: Both \gls{GCN} and GAT NEF perform worse, which we assume is because these models cannot take relative positions directly into account and instead only act on the existence or non-existence of edges.

At the same time, the introduction of fixed edge features to the \gls{GAT} model clearly shows its performance advantage, reducing the prediction error by a 30\% compared to the FF baseline.

We note that the comparatively bad performance of the \gls{IDM} in shorter timescales is consistent with previous work~\parencite{lenz_deep_2017} and it is likely to achieve better performances in a closed- or open-loop simulation.

\subsubsection{HighD}

On the HighD dataset, our results are different: As \figref{fig:highd_general_mae} and \tableref{tab:highd_performance} show, there is no significant performance difference between either of the learned models, and no significant performance difference between the \gls{IDM} and \gls{CVM} models. We believe this to be a consequence of little interaction between the cars, which makes all learned models degenerate to the non-interaction case and makes the interaction term of the \gls{CVM} model irrelevant. This shows that, even with no interaction, including interaction representations into our models does not cause performance degradation.

\vspace{\baselineskip}

In summary, we show that (A) several of our changes result in better performance, (B) as does a good interaction graph construction strategy. (C) In total, our model retains performance on a dataset with little interaction and greatly improves it on a dataset with plentiful interaction.

\section{Conclusion}
We have proposed modelling a traffic scene as a graph of interacting vehicles. Through this interpretation, we gain a flexible and abstract model for interactions.
To predict future traffic participant actions, we use \acrfullpl{GNN}, neural networks operating on graph data. These naturally take the graph model and therefore interaction into account. 
We evaluated two computationally efficient \glspl{GNN} and proposed several adaptations for our scenario.

In a traffic dataset with plentiful interaction, including interactions decreased prediction error by over 30\% compared to the best baseline model. At the same time, we saw no increase in prediction error on a dataset with little interaction.

While we have improved prediction quality, much work remains to be done: This work is only a proof-of-concept that modelling interactions as a graph is worthwhile and should thus be seen as only one technique for one aspect of traffic prediction. Integrating this model into existing state-of-the-art methodology, particularly \glspl{RNN}, remains an open task. At the same time, we would like to explore other graph construction strategies, particularly automatically finding relevant interactions.

\renewcommand*{\bibfont}{\footnotesize} %
\IEEEtriggeratref{15}
\printbibliography

\end{document}